\documentclass{bmvc2k}

\title{RED: Reinforced Encoder-Decoder Networks for Action Anticipation}

\addauthor{Jiyang Gao}{jiyangga@usc.edu}{1}
\addauthor{Zhenheng Yang}{zhenheny@usc.edu}{1}
\addauthor{Ram Nevatia}{nevatia@usc.edu}{1}

\addinstitution{
 Institute for Robotics and Intelligent Systems\\
 University of Southern California\\
 Los Angeles, CA, USA
}

\runninghead{}{RED: Reinforced Encoder-Decoder Networks for Action Anticipation}

\begin{document}

\maketitle

\begin{abstract}
Action anticipation aims to detect an action before it happens. Many real world applications in robotics and surveillance are related to this predictive capability. Current methods address this problem by first anticipating visual representations of future frames and then categorizing the anticipated representations to actions. However, anticipation is based on a single past frame's representation, which ignores the history trend. Besides, it can only anticipate a fixed future time. We propose a Reinforced Encoder-Decoder (RED) network for action anticipation. RED takes multiple history representations as input and learns to anticipate a sequence of future representations. One salient aspect of RED is that a reinforcement module is adopted to provide sequence-level supervision; the reward function is designed to encourage the system to make correct predictions as early as possible. We test RED on TVSeries, THUMOS-14 and TV-Human-Interaction datasets for action anticipation and achieve state-of-the-art performance on all datasets. 
\end{abstract}

\section{Introduction}

Action anticipation refers to detection (\emph{i.e. anticipation}) of an action before it happens. Many real world applications are related to this predictive capability, for example,  a surveillance system can raise alarm before an accident happens, and allow for intervention; robots can use anticipation of human actions to make better plans and interactions \cite{koppula2013anticipating}. Note that, online action detection \cite{de2016online} can be viewed as a special case for action anticipation, where the anticipation time is $0$.

Action anticipation is challenging for many reasons. First, it needs to overcome all the difficulties of action detection which require strong discriminative representations of video clips and ability to separate action instances from the large and wide variety of irrelevant background data. Then, for anticipation, the representation needs to capture sufficient historical and contextual information to make future predictions that are seconds ahead.

\begin{figure*}[h]
\centering
\includegraphics[scale=0.45]{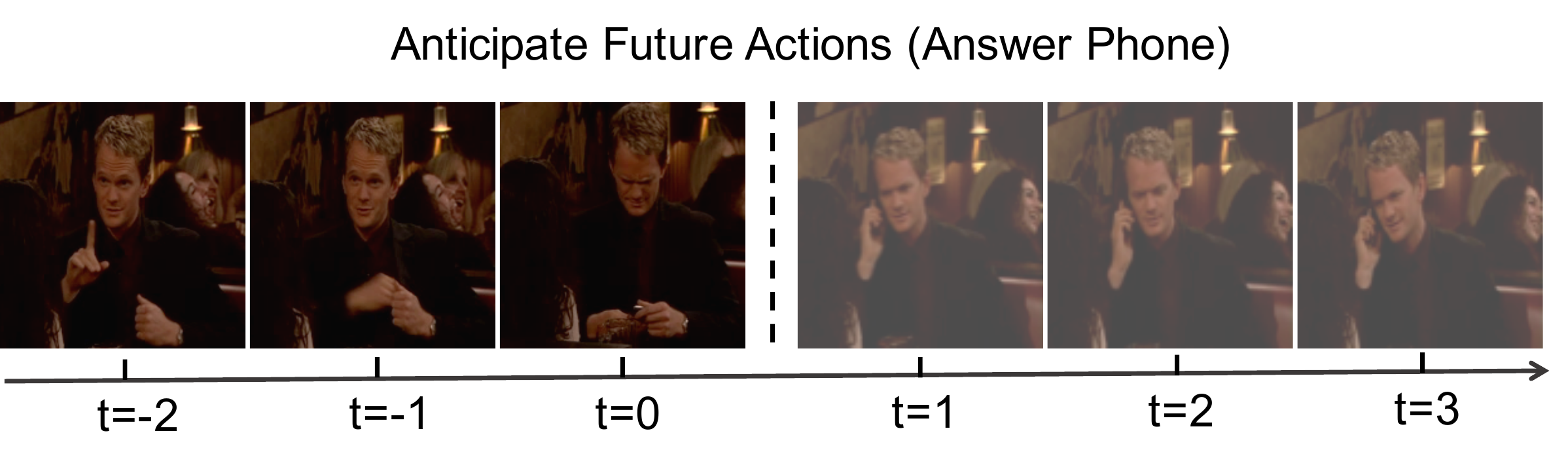}
\caption{Anticipating future actions by inferring from history information: the normal images represent past frames and the transparent images represent future frames. }
\end{figure*}

State-of-the-art methods on online action detection \cite{Ma_2016_CVPR, de2016online, yeung2015every} learn LSTM networks to encode history information and predict actions based on the hidden state of the LSTM. For action anticipation, early work \cite{lan2014hierarchical, pei2011parsing} was based on  traditional hand-crafted features. Recently, Vondrick \emph{et al.} \cite{vondrick2016anticipating} proposed to use deep neural networks to first anticipate visual representations of future frames and then categorize the anticipated representations to actions. 
However, the future representation is anticipated based on a single past frame's representation, while actions are better modeled in a clip, \emph{i.e.} multiple frames. Besides, their model only anticipates for single fixed time, it is desirable to be able to anticipate a sequence of continuous future representations. 

To address the anticipation challenges, we propose a Reinforced Encoder-Decoder (RED) network. The encoder-decoder network takes continuous steps of history visual representations as input and outputs a sequence of anticipated future representations. These anticipated representations are processed by a classification network for action classification. Squared loss is used for the representation anticipation and cross-entropy loss is used for action category anticipation (classification) during training. One drawback of the traditional cross-entropy loss is that it only optimizes the encoder-decoder networks greedily at each time step, and lacks sequence level optimization \cite{ranzato2015sequence}. We propose to use reinforcement learning to train the encoder-decoder networks on sequence level. The reward function is designed to encourage the model to make the correct anticipations as early as possible. We test RED on TVSeries  \cite{de2016online}, THUMOS-14 and TV-Human-Interaction \cite{patron2010high} for action anticipation and online action detection. State-of-the-art performance has been achieved.

\section{Related Work}

In this section, we introduce work on related topics, including online action detection, offline action detection, action anticipation and reinforcement learning in vision. 

\textbf{Early and Online Action Detection}
Hoai \emph{et al.} \cite{hoai2012max, hoai2014max} first proposed the problem of early event detection. They designed a max-margin framework which is based on structured output SVMs. Ma \emph{et al.} \cite{Ma_2016_CVPR}  address the problem of early action detection. They propose to train an LSTM network with ranking loss and merge the detection spans based on the frame-wise prediction scores generated by the LSTM. Recently, Geest \emph{et al.} \cite{de2016online} published a new dataset for online action detection, which consists of 16 hours (27 episodes) of TV series with temporal annotation for 30 action categories. 

\textbf{Offline Action Detection}
In the setting of offline action detection, the whole video is given and the task is to detect whether given actions occurs in this video and when does it occurs.  S-CNN \cite{Shou_2016_CVPR} presented a two-stage action localization framework: first using a proposal network to generate temporal proposals and then scoring the proposals with a localization network. TURN \cite{gao2017turn} proposed to use temporal coordinate regression to refine action boundaries for temporal proposal generation, which is proved to be effective and could be generalized to different action domains. TALL \cite{gao2017tall} used natural language as query to localize actions in long videos and designed a cross-modal regression model to solve it.

\textbf{Action Anticipation}
There have been some promising works on anticipating future action categories. Lan \emph{et al.} \cite{lan2014hierarchical} designed a hierarchical representation, which describes human movements at multiple levels of granularities, to predict future actions in the wild. Pei \emph{et al.} \cite{pei2011parsing} proposed an event parsing algorithm by using Stochastic Context Sensitive Grammar (SCSG) for inferring the goal of agents, and predicting the intended actions. Xie \emph{et al.} \cite{xie2013inferring} proposed to infer people's intention of performing actions, which is a good clue for predicting future actions. Vondrick \emph{et al.} \cite{vondrick2016anticipating} proposed to anticipate visual representation by training CNN on large-scale unlabelled video data.

\textbf{Reinforcement Learning in Vision}
We get inspiration from recent approaches that used REINFORCE \cite{williams1992simple} to learn task-specific policies. Yeung \emph{et al.} \cite{Yeung_2016_CVPR} proposed to learn policies to predict next observation location for action detection task by using LSTM networks. Mnih \emph{et al.} \cite{mnih2014recurrent} proposed to adaptively select a sequence of regions in images and only processing the selected regions at high resolution for the image classification task. Ranzato \emph{et al.} \cite{ranzato2015sequence} proposed a sequence-level training algorithm for image captioning that directly optimizes the metric used at test time by policy gradient methods.

\section{Reinforced Encoder-Decoder Network}
\begin{figure*}[]
\centering
\includegraphics[scale=0.47]{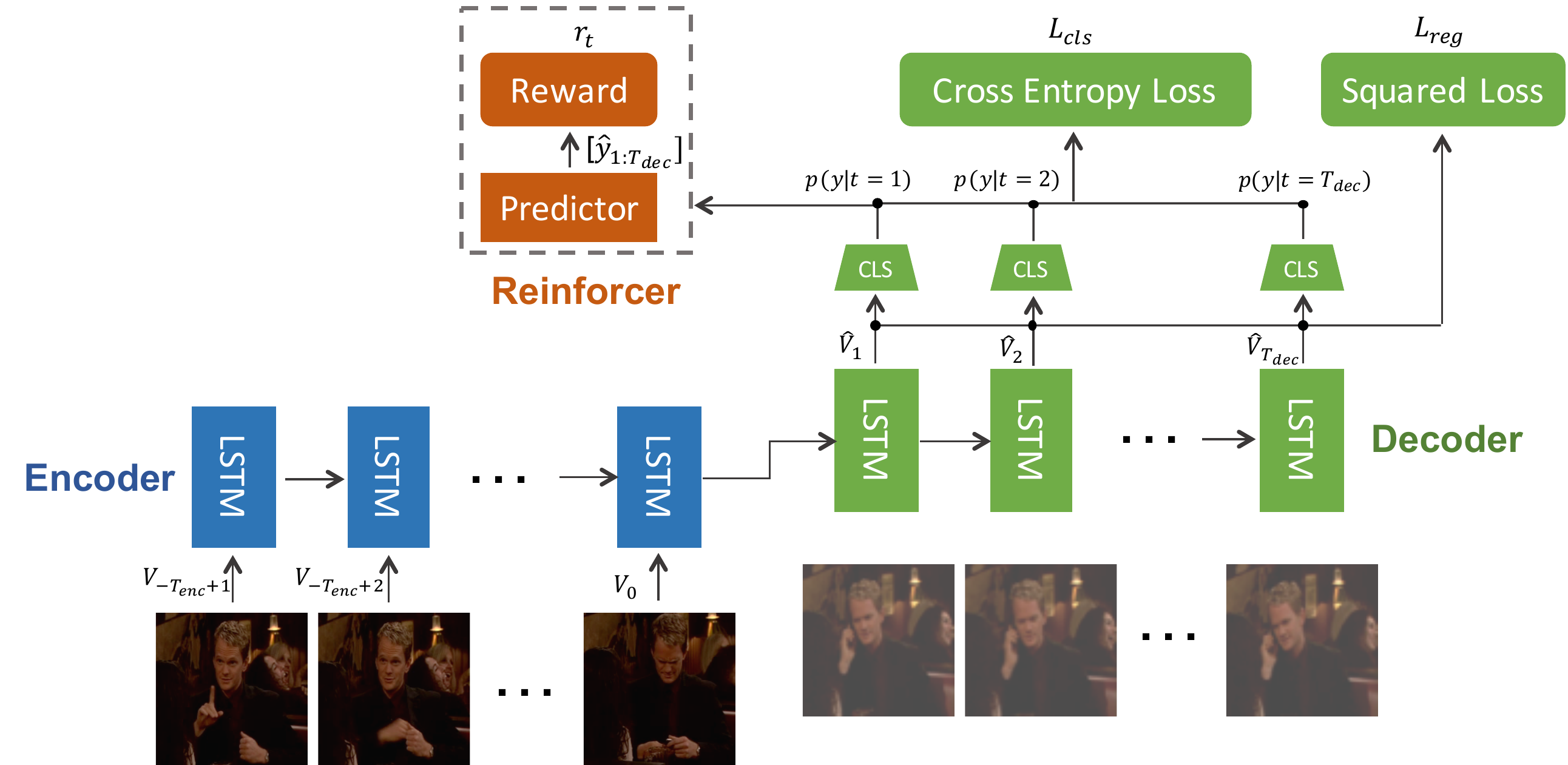}
\caption{Reinforced Encoder-Decoder (RED) networks architecture for action anticipation. }
\label{model}
\end{figure*}

RED contains three modules: a video representation extractor; an encoder-decoder network to encode history information and anticipate future video representations; a classification network to anticipate action categories and a reinforcement module to calculate rewards, which is incorporated in training phase using a policy gradient algorithm \cite{williams1992simple}. The architecture is shown in Figure \ref{model}.

\subsection{Video Processing}
A video is segmented into small chunks, each chunk contains $f=6$ consecutive frames. The video chunks are processed by a feature extractor $E_v$. It takes the video chunks $u_i$ as input and outputs chunk representation $V_i=E_v(u_i)$. More details on video pre-processing and feature extractors could be found in Section 4.1.

\subsection{Encoder-Decoder Network}
The encoder-decoder network uses a LSTM network as basic cell. The input to this network is a vector sequence $S_{in}=\{V_i\}, i \in [t-T_{enc},t)$, vector $V_i$ is a chunk visual representation, $T_{enc}$ is the length of the input sequence, $t$ is the time point in the video. After the last input vector has been read, the decoder LSTM takes over the last hidden state of encoder LSTM and outputs a prediction for the target sequence $S_{out}=\{\hat{V}_j\} j \in [t,t+T_{dec})$, where $T_{dec}$ is the length of the output sequence, \emph{i.e.}, the anticipation steps. The target sequence are representations of the video chunks that come after the input sequence.

The goal of decoder LSTM is to regress future visual representations, based on the last hidden state of the encoder networks. The loss function for training the encoder-decoder networks is the squared loss, 
\begin{equation}
L_{reg}=\frac{1}{N}\sum_{k=1}^{N}\sum_{j=1}^{T_{dec}}||\hat{V}_j^k-V_j^k||
\end{equation}
where $N$ is the batch size, $\hat{V}_j^k$ is the anticipated representation and $V_j^k$ is the ground truth representation. 

\subsection{Classification Network}
The output vector sequence, $S_{out}$, of the encoder-decoder networks is processed by the classification network, which has two fully connected layers, to output a classification distribution on action categories. The loss function of classification is the cross-entropy loss: $L_{cls}=\frac{1}{N}\sum_{k=1}^{N}\sum_{t=1}^{T_{dec}} log(p(y_t^k|y_{1:t-1}^k))$, 
where $p(y_k^j)$ is the probability score.

\subsection{Reinforcement Module}
A natural expectation of action anticipation is to make the correct anticipation as early as possible. For example, we consider two anticipation sequences "000111" and "001110", assuming the ground truth sequence is "011111", where "1" represents that an action category is happening and "0" represents that no action is happening (\emph{i.e} background). "001110" gives the correct anticipation earlier then "000111", so we consider it is a better anticipation at sequence level. However, cross-entropy loss would not capture such sequence-level distinctions, as it is calculated at each step to output higher confident scores on the ground truth category,  and no sequence-level information is involved.

To consider sequence-level reward, we incorporate reinforcement learning into our system. The anticipation module (the encoder-decoder networks) and the classification module (the FC layers) together can be viewed as an agent, which interacts with the external environment (the feature vector taken as input at every time step). The parameters of this agent define a policy, whose execution results in the agent making an \emph{prediction}. In the action detection and anticipation setting, a \emph{prediction} refers to predicting the action category in the sequence at each time step. After making a \emph{prediction}, the agent updates its internal state (the hidden state of LSTM), and observes a reward. 

We design a reward function to encourage the agent to make the correct anticipation as early as possible. Assuming the agent outputs an anticipation sequence $\{\hat{y}_i\}$, and the corresponding ground truth labels are $\{y_i\}$. In the ground truth label sequence, we denote the time position $t_f$ that the label start to change from background to some action class as \emph{transferring time}, for example, in "001111", $t_f=2$. At each step $t$ of the anticipation, the reward $r_t$ is calculated as 
\begin{equation}
r_t= \frac{\alpha}{t+1-t_f}, \text{ if  } t\geq t_f \text{ and } \hat{y}_t=y_t; ~~~ r_t=0, \text{  otherwise}
\end{equation}
where $\alpha$ is a constant parameter. If the agent makes the correct prediction at the transferring time $t_s$, it would receive the largest reward. The reward for making correct anticipation decays with time. The cumulative reward of the sequence is calculated as $R=\sum_{t=1}^{T_{dec}}r_t$.

The goal is to maximize the reward expectation $R$ when interacting with the environment, that is, to encourage the agent to output correct anticipations as early as possible. More formally, the policy of the agent induces a distribution over possible anticipation sequences $p_{y(1:T)}$, and we want to maximize the reward under this distribution:

\begin{equation}
J(\theta)=E_{p(y(1:t));\theta)}[\sum_{t=1}^{T_{dec}} r_t]=E_{p(y(1:t));\theta)}[R]
\end{equation}
where $y(1:t)$ is the action category predicted by our model. To maximize $J(\theta)$, as shown in REINFORCE\cite{williams1992simple}, an approximation to the gradient is given by 
\begin{equation}
\nabla_\theta J\approx\frac{1}{N}\sum_{k=1}^N\sum_{t=1}^{T_{dec}}\nabla_\theta \text{log}\pi(a_t^k|h_{1:t}^k,a_{1:t-1}^k)R_t^k
\end{equation}
where $\pi$ is the agent's policy. In our case, the policy $\pi$ is the probability distribution over action categories at each time step. Thus, the gradient can be written as
\begin{equation}
\nabla_\theta J\approx\frac{1}{N}\sum_{k=1}^N\sum_{t=1}^{T_{dec}}\nabla_\theta \text{log}p(y_t^k|y_{1:t-1}^k)(R_t^k-b_t^k)
\end{equation}
where $b_t^k$ is a baseline reward, which is estimated by a separate network. The network consists of two fully connected layer and takes the last hidden state of the encoder network as input. 

\subsection{Training Procedure}

The goal of encoder-decoder networks is to anticipate future representations, so unlabelled video segments (no matter whether there contain actions) could be used as training samples. As the positive segments (\emph{i.e.} some action is happening) are very small part of videos in the whole datasets, RED is trained by a two-stage process. In the first stage, the encoder-decoder networks for representation anticipation are trained by the regression loss $L_{reg}$ on all training videos in the dataset as initialization. In the second stage, the encoder-decoder networks are optimized by the overall loss function $L$, which includes the regression loss $L_{reg}$, a cross-entropy loss $L_{cls}$ introduced by classification networks and $J$ introduced from reinforcement module on the positive samples in the videos: 
\begin{equation}
L=L_{reg}+L_{cls}-J
\end{equation}
The classification network is only trained in the second stage by $L_{cls}$ and $J$. 

The training samples for the first stage do not require any annotations, so they could be collected at any position in the videos. Specifically, at a time point $t$, the sequence for encoder networks is $[V_{t-T_{enc}},V_{t})$, the output ground truth sequence for decoder networks is $[V_{t},V_{t+T_{dec}})$, where $V_t$ is the visual representation at $t$. For the second stage, the training samples are collected around positive action intervals. Specifically, given a positive action interval $[t_s, t_e]$, the central time point $t$ could be selected from $t>t_s-T_{enc}$ to $t<t_e$. After picking the central time point $t$, $[V_{t-T_{enc}},V_{t})$ are used as input samples of the encoder, $[V_{t},V_{t+T_{dec}})$ are used as output ground truth visual representations for the anticipated sequence. $[y_{t},y_{t+T_{dec}})$ are ground truth action labels for the anticipated sequence, which are used in classification cross-entropy loss.

\section{Evaluation}

We evaluate our approach on standard benchmarks for action anticipation, including TVSeries \cite{de2016online}, THUMOS-14 and TV-Human-Interaction \cite{patron2010high}. As no previous action anticipation results are available on THUMOS-14 and TVSeries, we build several strong baselines (including one similar to \cite{vondrick2016anticipating}) to compare with. 
\subsection{Implementation Details}

We extract frames from all videos at 24 Frames Per Second (FPS). The video chunk size is set to 6, so each chunk is 0.25-second long in the original video. We investigate two video feature extractors: a two-stream CNN model \cite{xiong2016cuhk} and VGG-16 \cite{simonyan2014very} model. For two-stream model, we use central frame to calculate the appearance CNN feature in each chunk. The outputs of  "Flatten\_673" layer in ResNet are extracted;  we calculate the optical flows \cite{farneback2003two} between the $6$ consecutive frames in a chunk, and feed them into the motion stream. The output of "global pool" layer in BN-Inception \cite{ioffe2015batch} is used. We concatenate the motion features and the appearance features into 4096-dimensional vectors, which is used as chunk representation. To provide fair comparison with \cite{de2016online}, we also use VGG-16 features. The central frame in a chunk is sampled and processed by a VGG-16 model (pre-trained on ImageNet), the output of $fc$6 is used as the chunk representation. 

We set $T_{enc}=16$ and $T_{dec}=8$, which are 4 seconds and 2 seconds respectively. The hidden state of LSTM is set to 4096-dimensional and number of LSTM layer is 1. We also tried 2048-dimensional, but shows a worse performance. $\alpha$ in Equation (2) is set to 1. Adam \cite{kingma2014adam} is adopted to train RED, learning rate is set to 0.001, and batch size is 64.

\subsection{Baseline Methods}

\begin{figure*}[h]
\centering
\includegraphics[scale=0.43]{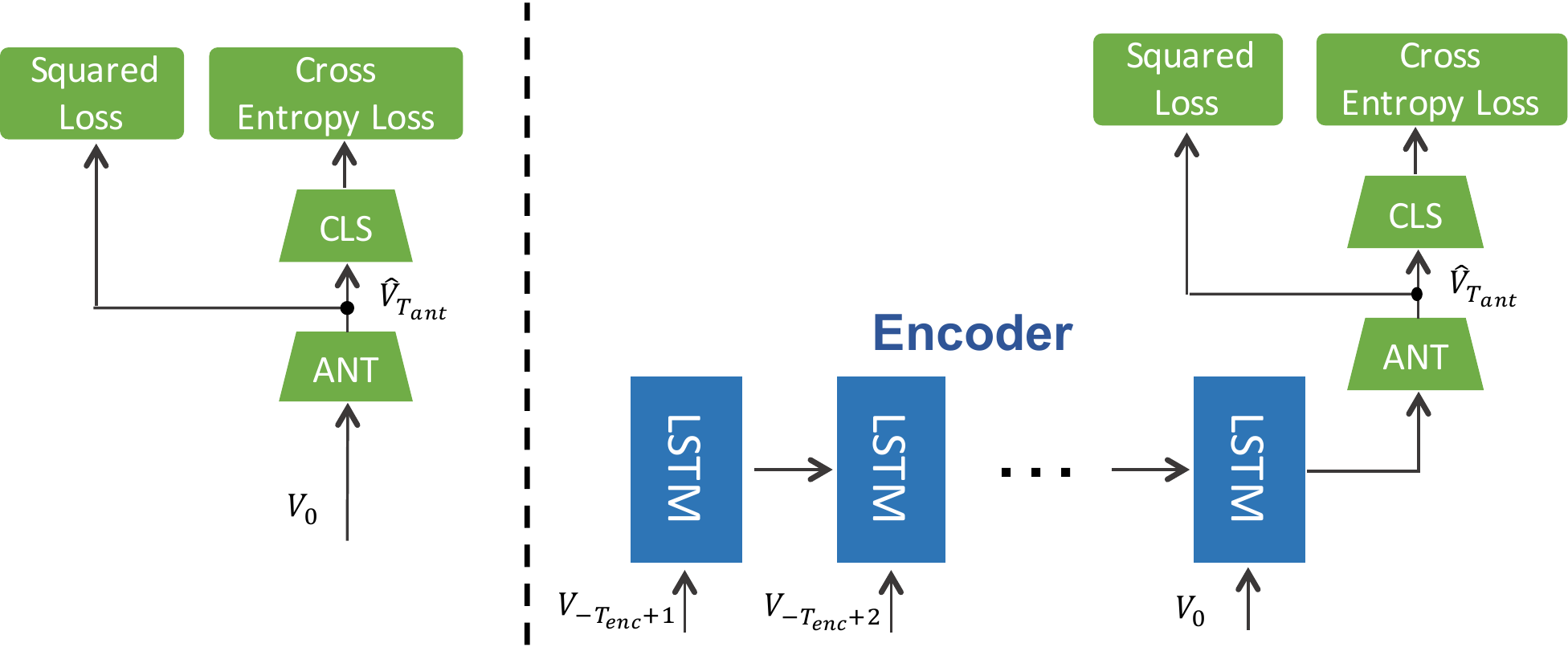}
\caption{Baseline methods for action anticipation. The left one is FC and the right one is EFC.}
\label{baseline}
\end{figure*}

\textbf{FC}: We implement a method similar to that in \cite{vondrick2016anticipating}, in which fully connected layers are trained to anticipate the future representation, as shown in Figure \ref{baseline}, on the left. Specifically, the chunk representation at time $t$ are input to two fully connected layers and regressed to the future chunk representations at time $t+T_{ant}$. The anticipating time is $T_{ant}$. The anticipated representations are processed by another classification network, which consists of two fully connected layers. The outputs of the classification network are the action category anticipations. $T_{ant}$ is set to 4, which is 1 second.

\textbf{Encoder-FC (EFC)}: The above baseline method \textbf{FC} only considers a single representation for anticipation. This baseline method extends to fully connected layers to a LSTM encoder network, which encodes the representations from $t-T_{enc}$ to $t$, and  anticipates the representation at $t+T_{ant}$, as shown in Figure \ref{baseline}, on the right. The classification layers take the anticipated representations to anticipate action categories. We set $T_{enc}=16$, the same as the one of RED. $T_{ant}$ is set to 4.

\textbf{Encoder-Decoder (ED)}: This baseline is similar to RED, but without the reinforcement module. The EFC baseline method considers sequenced history representations and output a single anticipated video representation at time $t+T_{ant}$. Instead of anticipating a single future representation, the decoder learns to anticipate multiple continuous representations from time $t$ to $t+T_{dec}$. We set $T_{enc}=16$ and $T_{dec}=8$, the same as the ones of RED.

The aforementioned baseline models are trained by a two-stage process, which is similar to RED. In the first stage, the representation anticipation networks, \emph{i.e.} 2-layer FC anticipation, EFC anticipation or encoder-decoder anticipation, are trained by a regression loss on all the videos in the dataset (THUMOS-14 or TVSeries) as initialization. In the second stage, the anticipation networks are trained by the regression loss and a cross-entropy loss, which is introduced by the classification network on the positive samples in the videos. 

\subsection {Experiments}

\textbf{Dataset.} The TVSeries Dataset \cite{de2016online} is a realistic, large-scale dataset for temporal action detection; it contains 16 hours of videos (27 episodes) from six recent popular TV series. TVSeries contains 30 daily life action categories, such as "answer phone", "drive car", "close door". 
The temporal detection part of THUMOS-14 contains over 20 hours of videos from 20 sport classes. The training set of THUMOS-14 contains only trimmed videos, which are not suitable for action anticipation and action detection. There are 200 and 213 untrimmed  videos in the validation and test set respectively. We train RED on the validation set and test it on the test set.
TV-Human-Interaction \cite{patron2010high} dataset contains 300 video clips extracted from 23 different TV shows. There are 4 interactions: hand shakes, high fives, hugs and kisses in the dataset. The video lengths range from 1 second to 10 seconds.

\textbf{Experiment setup.} Training representation anticipation (stage 1) needs tens of hours of video data. TVSeries and THUMOS-14 have enough videos, but TV-Human-Interaction only contains tens of minutes. So we train and test our model on both TVSeries and THUMOS-14, and compare the performance of RED for anticipation time $T_a=0.25s-2.0s$. For TV-Human-Interaction, we use the stage-1 model trained on THUMOS-14 or TVSeries, and train stage-2 on TV-Human-Interaction.

On TVSeries, the metric we use is calibrated Average Precision (cAP) which is proposed by \cite{de2016online}. Calibrated precision (cPrec) uses a parameter $w$, which the ratio between negative frames and positive frames in the calculation of precision, $cPrec=\frac{TP}{TP+FP/w}$, so that the average precision is calculated as if there were an equal amount of positive and negative frames. On THUMOS-14, we report per-frame mean Average Prevision (mAP) performance, which is used in \cite{yeung2015every}. For TV-Human-Interaction, we report classification accuracy (ACC).

\begin{table}[h]
\centering
\caption{Action anticipation comparison on TVSeries (cAP \%) test set and THUMOS-14 (per-frame mAP \%) test set at 1s (4 chunks) with two-stream features. }
\label{tv-ta1s}
\begin{tabular}{l|cccc}
\hline
    & FC & EFC & ED & RED \\ \hline
TVSeries (cAP@$T_a$=1s \%) &  72.4  &   73.3   &  74.6  &  \textbf{75.5}   \\ \hline
THUMOS-14 (mAP@$T_a$=1s \%) &  31.7  &   33.9   &  36.8  &  \textbf{37.5}   \\ \hline
\end{tabular}
\end{table}

\textbf{Comparison of action anticipation methods.}
We first compare the performance at anticipation time $T_a=1s$. The results on TVSeries and THUMOS-14 are shown in Table \ref{tv-ta1s}. Overall, RED outperforms FC on both datasets. Comparing with FC and EFC, we can see that encoding multiple history steps of representation improves the anticipation performance consistently. 
The difference between ED and EFC is the presence of  decoder networks, EFC uses fully connected layers to anticipate only one future representation, but ED uses decoder networks to anticipate a sequence of continuous representations. Comparing ED and EFC, we can see that anticipating the future representation step by step makes the anticipations more accurate. Comparing ED and RED, it can be seen that the proposed reinforer benefits action anticipation. 

\begin{table}[h]\footnotesize
\centering
\caption{Action anticipation comparison (ACC \%) on TV-Human-Interaction at $T_a=1s$ (4 chunks).}
\label{interaction-ta1s}
\begin{tabular}{l|ccc}
\hline
    & Vondrick \emph{et al.}\cite{vondrick2016anticipating} (THUMOS) & RED-VGG (TVSeries) & RED-TS (THUMOS) \\ \hline
ACC@$T_a$=1s (\%) &  43.6    &  47.5  &  \textbf{50.2}   \\ \hline
\end{tabular}
\end{table}

We compare RED with Vondrick \emph{et al.} \cite{vondrick2016anticipating} on TV-Human-Interaction at anticipation time $T_a=1s$, as shown in Table \ref{interaction-ta1s}. Stage-1 of RED is trained on TVSeries (with VGG feature) or THUMOS-14 (with two stream feature), stage-2 is trained on TV-Human-Interaction. Note that, Vondrick \emph{et al.} \cite{vondrick2016anticipating} trained the representation anticipation model on THUMOS dataset and trained SVM classifiers for action anticipation. We can see that both RED-VGG and RED-TS outperform \cite{vondrick2016anticipating}. Note that \cite{vondrick2016anticipating} uses Alexnet features so possibly some of our gain comes just from use of stronger features. On the other hand, we use a much smaller subset of THUMOS (20 hours vs 400 hours) to train. Also, Table \ref{tv-ta1s}, which includes a comparison with our implementation of \cite{vondrick2016anticipating},  and using the same TS features as RED,  does indicate that our network design is responsible for significant part of the improvements.

\begin{table}[h]
\centering
\caption{Detailed action anticipation (cAP \%) comparison for ED and RED on TVSeries test set from $T_a=0.25s$ to $T_a=2.0s$ with two-stream representations and VGG features.}
\label{tv-ed-red}
\begin{tabular}{l|llllllll}
\hline
time &0.25s & 0.5s & 0.75s & 1.0s & 1.25s & 1.5s & 1.75s & 2.0s \\ \hline
ED-VGG   &  71.0 & 70.6 & 69.9 & 68.8    & 68.0     & 67.4   &     67.0 &  66.7   \\ 
RED-VGG     & \textbf{71.2} &  \textbf{71.0}   &\textbf{70.6}  & \textbf{70.2}&  \textbf{69.2}    &    \textbf{68.5} & \textbf{67.5}     &  \textbf{66.8}  \\ \hline

ED-TS   &  78.5 & 78.0 & 76.3     & 74.6    & 73.7     & 72.7   &     71.7 &  71.0   \\ 
RED-TS   & \textbf{79.2} &  \textbf{78.7}   &\textbf{77.1}  & \textbf{75.5}&  \textbf{74.2}    &    \textbf{73.0} & \textbf{72.0}     &  \textbf{71.2}  \\ \hline
\end{tabular}
\end{table}

\begin{table}[h]
\centering
\caption{Detailed action anticipation (per-frame mAP \%) comparison for ED and RED on THUMOS-14 test set from $T_a=0.25s$ to $T_a=2s$ with two-stream representations.}
\label{thumos-ed-red}
\begin{tabular}{l|llllllll}
\hline
time  &0.25s & 0.5s & 0.75s & 1.0s & 1.25s & 1.5s & 1.75s & 2.0s \\ \hline

ED-TS     & 43.8  & 40.9 &   38.7   &   36.8  &  34.6  & 33.9  & 32.5 & 31.6   \\ 
RED-TS    & \textbf{45.3} &  \textbf{42.1}   &\textbf{39.6}  & \textbf{37.5}&  \textbf{35.8}    &    \textbf{34.4} & \textbf{33.2}     &  \textbf{32.1}  \\ \hline
\end{tabular}
\end{table}

\textbf{Varying anticipation time.} Encoder-decoder network allows for sequence anticipation, unlike FC or EFC which can only anticipate at a fixed future time. A detailed comparison between "ED" and "RED" is shown in Table \ref{tv-ed-red} and Table \ref{thumos-ed-red}. We test RED with two types of video representations on TVSeries: two-stream model and VGG-16. On THUMOS-14, we report two-stream performance. The anticipation time ranges from 0.25s to 2s, which correspond to 1 video chunk and 8 video chunks (each chunk contains 6 frames, and frame extraction rate is 24). "TS" stands for two-stream. As shown in Table \ref{tv-ed-red}, it can be seen that reinforcement module consistently improves the encoder-decoder networks for action anticipation at each step on both two stream features and VGG features. We think it is because the sequence-level supervision introduced by the designed reward function makes the optimization more effective. The reward function together with the cross-entropy loss not only trains the system to make the correct action anticipation at each single step, but also encourages it to produce the correct sequence prediction as early as possible. The results on THUMOS-14 are consitent with those on TVSeries, the sequence-level optimization from reinforcement learning consistently benefits the action anticipation.

\textbf{Comparison of online action detection}
As discussed before, online action detection can be treated as a special case for action anticipation, where the anticipation time $T_a=0$. We use the results at minimum anticipation time ($T_a=0.25s$) for online action detection. As shown in Table \ref{tv-online}, the previous state-of-the-art methods based on CNN, LSTM and Fisher Vector (FV) in \cite{de2016online} achieved 60.8, 64.1 and 74.3 respectively. \cite{de2016online} uses VGG features for "CNN" and "LSTM" methods, which are same as our RED-VGG. RED-VGG achieves 71.8, outperforms 64.1 by a large margin. RED-TS achieves the highest performance 79.2. Similar to TVSeries, we use the results of $T_a=0.25s$ as results of online action detection on THUMOS-14. The results are shown in Table \ref{thumos-online} and we can see that RED outperforms state-of-the-art methods.

\begin{table}[h]
\centering
\caption{Comparison on online action detection in TVSeries test set. }
\label{tv-online}
\begin{tabular}{l|ccc|cc}
\hline
    & CNN\cite{de2016online} & LSTM\cite{de2016online} &FV \cite{de2016online} & RED-VGG & RED-TS \\ \hline
cAP (\%) &  60.8  &   64.1   &  74.3  &  71.2 &\textbf{79.2}   \\ \hline
\end{tabular}
\end{table}

\begin{table}[h]
\centering
\caption{Online action detection comparison on THUMOS-14 test set (per-frame mAP \%) with two stream features. }
\label{thumos-online}
\begin{tabular}{l|cccc}
\hline
    & two-stream\cite{simonyan2014two} & LSTM\cite{yeung2015every} & MultiLSTM\cite{yeung2015every} & RED \\ \hline
mAP(\%) &  36.2  &   39.3  &  41.3  &  \textbf{45.3}   \\ \hline
\end{tabular}
\end{table}


\section{Conclusion}
We propose Reinforced Encoder-Decoder (RED) networks for action anticipation. RED takes multiple history representations as input and learns to anticipate a sequence of future representations, which are fed into classification networks to categorize actions. The salient aspect of RED is that a reinforcement module is adopted to provide sequence-level supervision. The reward function is designed to encourage the system to make correct prediction as early as possible. RED is jointly optimized by the cross-entropy loss, squared loss and the reward function via a two-stage training process. Experimental results on action anticipation show the effectiveness of proposed reinforcement module and the encoder-decoder network. 

\bibliography{egbib}
\end{document}